\documentclass[letterpaper, 10 pt, conference]{ieeeconf}  

\IEEEoverridecommandlockouts                              

\usepackage{graphics} 
\usepackage{epsfig} 
\usepackage{mathptmx} 
\usepackage{times} 
\usepackage{amsmath} 
\usepackage{amssymb}  

\usepackage{xcolor}
\usepackage{graphicx}
\usepackage{enumerate}
\usepackage{algorithm}
\usepackage{algorithmic}
\usepackage{subcaption} 
\usepackage{wrapfig}
\usepackage{booktabs} 
\newcommand{\learnedCost}{C_\phi}
\newcommand{\irlloss}{\mathcal{L}_{IRL}}
\newcommand{\com}[1]{{\textcolor[rgb]{0.502, 0.502, 0.502}{#1}}}

\makeatletter
\newcommand\fs@norules{\def\@fs@cfont{\bfseries}\let\@fs@capt\floatc@ruled
  \def\@fs@pre{}%
  \def\@fs@post{}%
  \def\@fs@mid{\kern3pt}%
  \let\@fs@iftopcapt\iftrue}
\makeatother
\floatstyle{norules}
\restylefloat{algorithm}

\title{\LARGE \bf
Learning Time-Invariant Reward Functions through \\ Model-Based Inverse Reinforcement Learning
}

\author{Todor Davchev$^{\dagger \ast}$ Sarah Bechtle$^{\ddagger}$ Subramanian Ramamoorthy$^{\dagger}$ and Franziska Meier$^{\diamond}$
\thanks{$^{\ast}$ Corresponding Author.}%
\thanks{$^{\dagger}$ Authors are with the School of Informatics, University of Edinburgh, 10 Crichton St, EH8 9AB, United Kingdom, {\tt\small \{t.b.davchev, s.ramamoorthy\}@ed.ac.uk}}%

\thanks{$^{\ddagger}$ Author is with MPI for Intelligent Systems, Tubingen, Germany {\tt\small sbechtle@tuebingen.mpg.de}}%
\thanks{$^{\diamond}$ Author is with Facebook AI Research, Menlo Park, CA
{\tt\small fmeier@fb.com}}
}

\begin{document}

\maketitle
\thispagestyle{empty}
\pagestyle{empty}

\begin{abstract}

Inverse reinforcement learning is a paradigm motivated by the goal of learning general reward functions from demonstrated behaviours. Yet the notion of generality for learnt costs is often evaluated in terms of robustness to various spatial perturbations only, assuming deployment at fixed speeds of execution. However, this is impractical in the context of robotics and building, time-invariant solutions is of crucial importance. In this work, we propose a formulation that allows us to 1) vary the length of execution by learning time-invariant costs, and 2) relax the temporal alignment requirements for learning from demonstration. We apply our method to two different types of cost formulations and evaluate their performance in the context of learning reward functions for simulated placement and peg in hole tasks executed on a 7DoF Kuka IIWA arm. Our results show that our approach enables learning temporally invariant rewards from misaligned demonstration that can also generalise spatially to out of distribution tasks. 

\end{abstract}

\section{INTRODUCTION}

Learning from demonstration (LfD) bootstraps the process of learning new skills by utilising expert's knowledge. This is particularly useful for robot learning and object manipulation where sample efficiency is central. One established way of skill learning via demonstrations is to first infer the expert's intent through extracting a reward function that the robot can then use to learn a policy that matches the demonstrated skill. This enables the effective compression of complex skills and enables the robot to learn successful policies for various task-settings and environments. This can be useful for example when there is an insufficient number of demonstrations to recover the task distribution and therefore the policy directly as alternatively done by behaviour cloning \cite{osa2018algorithmic}.

Collecting demonstrations on physical robots at different times of the day or by using different experts can often result in sets of misaligned demonstrations both in terms of speed of motion and time duration. This can affect the performance of the learning algorithms we build. It can also be impractical as, for instance, we might want a robot to match a particular speed of a collaborator in an object handling task or for safety reasons in complex manipulation like surgery. Therefore building solutions that are both invariant to the time duration of different demonstrations and that can also scale to different speeds of execution without needing new sets of demonstrations is important. In this work, we address this problem. We propose a time-invariant reward learning approach that 1) can directly learn from temporally misaligned demonstrations with no required preprocessing; and 2) is able to deploy the task in a few-shot manner at various speeds of execution.

The problem of how to learn such rewards from demonstrations is often referred to as inverse reinforcement learning (IRL) \cite{abbeel2004apprenticeship, ziebart2008maximum, das2020model} or inverse optimal control (IOC) \cite{finn2016guidedirl, englert2017inverseRL}. Prior work on inverse reinforcement learning does not address the problem of learning reward functions that are invariant to time, duration and speed of motion, meaning reward functions need to be learned from demonstrations that are consistent along those variables. Once learnt, those rewards can be used only for fixed speed deployment. Yet, the process of collecting demonstrations in the physical world is often tedious and challenging, especially when they need to be temporally aligned e.g. for autonomous helicopter flying \cite{abbeel2010autonomous} or surgical robotics \cite{van2010superhuman, osa2017online}. As a result, a number of time warping solutions have been proposed that ensure demonstrations get aligned as a post processing step \cite{abbeel2010autonomous, van2010superhuman, osa2017online, jin2020learning}. However, this itself removes the problem of alignment from the learning objective completely and thus prevents from scaling the learnt rewards to different speeds of execution. Once extracted, those reward functions result in fixed policy deployment and cannot generalise to different speeds of execution. This is limiting since varying the speed of an execution can be of crucial importance to enabling slower and smoother or faster and more nimble executions that can for example match the speed of a collaborator or perform more careful, safer control.

\begin{figure*}
      \centering
      \includegraphics[width=\textwidth]{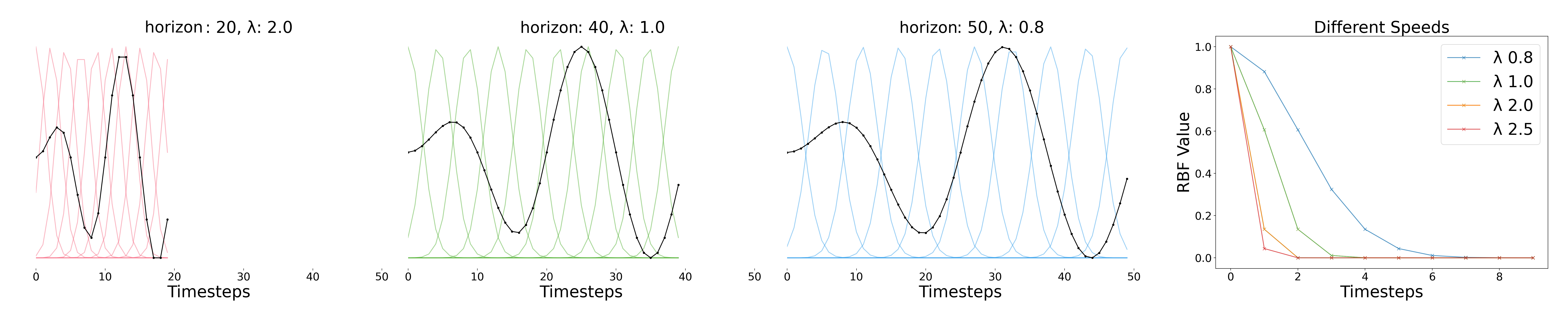}
      \caption{\small{Example RBF spacing for different speeds on a 1D sine motion. Left plot depicts the fastest execution which has fewer time steps and thus a shorter horizon. Assuming the same goal, a longer horizon will represent a slower execution. Colours indicate ways to temporally scale the ten centres of a weighted RBF cost function by relying on time-variable $\lambda$. The number of weights stays fixed while RBF centres shift in time staying the same relative to the sine curve. Final plot illustrates different RBF widths for a single centre at 0 and using varied temporal scaling and linear spacing. }
      }
      \label{fig:perturbations}
      \vspace{-5mm}
\end{figure*}

The focus of this work is to relax this assumption. We build upon recent advances on model-based IRL through bi-level optimisation \cite{das2020model} and propose a method for obtaining time-invariant rewards in the context of model-based IRL from misaligned demonstrations. Our solution relies on the intuition that particular weights can correspond to particular spatial regions in a given execution. We hypothesise that by keeping this relationship static while demonstrations shift in time (See Figure~\ref{fig:perturbations}) allows us to achieve time-invariance that helps with generalisation. We combine this notion with the ability of bi-level optimisation to enable fast and stable optimisation in multi-task settings. We propose two different ways for utilising our temporal scaling mechanism in the form of novel structured and unstructured time-invariant rewards. We show that our rewards can synthesise policies which can run at arbitrary speeds of execution and can be applied to a range of never seen before goals.

We evaluate our solution in simulation on two manipulation tasks, namely object placement and peg insertion, executed by a 7DoF Kuka IIWA arm. Concretely, the contributions of this work are:
1) We propose a temporal scaling mechanism for learning both time-invariant structured (linear in weights) and unstructured (neural networks) reward functions.
2) demonstrate that this formulation enables learning reward functions from temporally misaligned demonstrations that can be deployed at different speeds;
3) extensively compare the generalisation abilities of both structured and unstructured reward functions at test time with respect to speed and unseen goal locations, when learned with and without time-invariance.

Our study shows that our time-invariant rewards can be learned from multiple misaligned demonstrations. We also show how the proposed approach can generalise both to different goal locations and to different execution speeds. This is important in different settings like matching human collaborator's speed or for manipulating fragile objects that require slower manipulation speeds than sturdy ones.
\section{LITERATURE REVIEW}
\subsection{Model-free and Model-based IRL}
The inverse reinforcement learning field can be roughly split into 1) model-free \cite{ziebart2008maximum,boularias2011relativeirl} and 2) model-based \cite{abbeel2004apprenticeship, abbeel2010autonomous, das2020model} reward learning algorithms. Similarly to the model-based vs model-free reinforcement learning, model-based IRL is expected to be more sample-efficient but comes with the challenge of having to learn a model that captures the system dynamics.
Model-free IRL algorithms have previously shown to be useful in the context of robotics applications \cite{kalakrishnan_2013_irl, boularias2011relativeirl, finn2016guidedirl}. These approaches were shown to extract successful rewards when provided with careful feature selection. However, neither model-based nor model-free IRL methods have considered learning time-invariant reward functions. In this work, we propose a model-based IRL algorithm that learns time-invariant reward functions that are applicable to different speeds of execution by relying on bi-level optimisation.

Recent advances in using generative adversarial learning have shown  potential to learning rewards that can tackle complex tasks \cite{ho2016generative, liu2019state} and learn from incomplete demonstrations too \cite{sun2019adversarial}. However, those do not consider deployment at different speeds of execution, learning from misaligned demonstrations and do not target contexts of sample efficient, few-shot reward learning. This prevents them from being useful in practical robotics settings. In contrast, our solution does not assume access to large number of demonstrations (we use$\leq12$ in this work) and results in costs that can acquire successful policies in only a few update steps (1 and 5).

Most model-based IRL methods were built assuming pre-processed, temporally aligned demonstrations. Those algorithms have previously represented costs formulated as linear mappings between learnable weights and spatially relevant features for simulated tasks \cite{abbeel2004apprenticeship, ziebart2008maximum, abbeel2010autonomous}, or structured, time-dependent weights with squared features for robotics tasks \cite{englert2017inverseRL, das2020model}. Assuming linear dependency in the weights is generally a great approach as it leads to intuitive and explainable costs that are ideal for running on physical systems. However, they often rely on expert engineering which may be difficult to obtain and may limit an algorithm's capacity to generalise. Some recent works have also considered using unstructured cost function representations \cite{finn2016guidedirl, zhaoinverse} which rely a lot less on expert feature engineering and thus hold a good potential for extracting more general features. However, both linear and nonlinear solutions have never been shown to be robust to learning with temporally misaligned demonstrations or to generalise to various speeds of execution. 
\subsection{Temporal Invariance in Learning from Demonstration}
There have been a range of LfD methods that have considered adding temporal invariance to learned behaviours. Those can be roughly split into two categories, approaches that learn policies directly from demonstrations, and ones that infer rewards from demonstrations.

\cite{khansari2011learning, khansari2014learning, gribovskaya2011learning, ijspeert2013dynamical, koutras2020dynamic} learn stable, time-invariant dynamical systems by utilising a temporal scaling variable similar to the one proposed in this work. However, those behavioural cloning (BC) based works infer policies from demonstrations directly. Although valid approaches, they rely on sufficient number of demonstrations that can effectively recover the distribution that describes the task at hand \cite{osa2018algorithmic}. In contrast, we assume an insufficient numbers of demonstrations that require inferring the experts intent by recovering a reward function. The problem of inferring time-invariant rewards, which are then used to synthesise successful policies, however is largely unexplored. \cite{noseworthy2020task, hristov2020learning} consider learning latent policy representations from multiple demonstrations. They focus on disentangling different manners of executions in a learnt latent space that enable executions with different speeds. Those works, however can only scale to the speeds present in the training data and are known to require large number of demonstrations.
Recent work, \cite{jin2020learning}, proposes a gradient-based solution from sparse demonstrations by learning a warping function capable of adapting demonstrations to feasible timescales. However, their focus is on scaling the demonstrator's trajectories to the capabilities of a particular robot as opposed to enabling different speeds of execution. Further, they do not address the problem of learning from multiple misaligned demonstrations.
\subsection{Summary}
In summary, we propose a method for learning temporally-invariant costs for model-based inverse reinforcement learning. We demonstrate the utility of the proposed two costs to scale to different speeds of execution by relaxing the requirements for temporally aligned demonstrations. We evaluate our work in the context of lifelong learning against different simulated manipulation tasks.

\section{MODEL-BASED INVERSE REINFORCEMENT LEARNING WITH TEMPORAL SCALING}
In this work we learn time invariant cost functions, $C_{\phi}$ from misaligned demonstrations $\tau_{demo}$ and use them to synthesise policies $\pi_{c_{\phi}}$ that can generalise to different temporal and spatial perturbations.  We build upon a recently proposed model-based IRL approach \cite{das2020model} that relies on advances of bi-level optimisation \cite{bechtle2021meta,higher}. This approach consists of two nested loops - an outer loop responsible for updating the current cost's parameters $\phi$, and an inner loop used to optimise $\pi_{c_{\phi}}$ using model-predictive control (MPC) with the current cost $C_{\phi}$ and parameters $\phi$
(see Alg.~\ref{algo:cost-learning-meta-train-individual}, lines 4 and 10). Such a formulation is both stable and sample efficient due to its model based nature. It can also synthesise successful policies in a few-shot learning manner thanks to bi-level optimisation. However, as it stands this formulation does not necessarily allow for varying the speed of execution of the extracted policies. This can be limiting both in terms of practicality of the solution but also can affect its spatial generality. We propose a temporal scaling mechanism for learning time-invariant costs that can be learnt from temporally misaligned demonstrations that addresses this problem. 

\subsection{Problem Formulation}
We consider deterministic, fixed-horizon and discrete time control tasks with continuous states $\mathbf{s} = (s_1, \dots, s_T)$ and continuous actions $\mathbf{u} = (u_1, . . . , u_T)$. Each state
$s_t$ is comprised of a latent representation $z_t$ at some time step $t$. The control tasks are characterized by a differentiable dynamics model \cite{sutanto2020encoding}, $\hat{s}_{t+1} = f_{dyn}(s_t, u_t)$ and a cost function $\learnedCost$. The objective then is to find the optimal parameters $\phi$ for $C_{\phi}(\cdot)$ that effectively summarises some strategy for performing a control task. We learn a policy $\pi$ using $C_{\phi}(\cdot)$ as an optimisation objective. We can then define an optimal control policy as $\pi^{*}=\arg\min_{\pi \in \bar{\pi}}J(\pi)$, where $J(\pi) = \frac{1}{N}\sum^{t+N}_{\tau=t+1}C_{\phi}(\hat{s}_t, g)$ with some goal $g$. Finally, we use gradient decent to extract optimal control actions by unrolling an entire trajectory (rows 10-15, Alg.~\ref{algo:cost-learning-meta-train-individual}). In this work, we propose to learn costs that work for trajectories of varying lengths.
\begin{algorithm}[H]
\begin{algorithmic}[1]
\footnotesize{
 \renewcommand{\algorithmicrequire}{\textbf{Input:}}
 \renewcommand{\algorithmicensure}{\textbf{Output:}}
\REQUIRE Initial $\phi$, learning rates $\eta=.001, \alpha=.01$, demos $\tau_{demo}$, \\ base duration $|\tau_{base}|$ and a fixed linearly spaced variable $x$ \\ initial states $s_0 = (\theta_0, \dot{\theta}_0, z_0)$
\ENSURE  Learnt $\phi$
\FOR{each $epoch$}
\FOR{each demo}
\STATE{$u_t = 0, \forall t=1,\dots, T_{demo}$}
\STATE{assign new goal state $g = \hat{\tau}_{T_{demo}}$}
\STATE{extract demo duration $|\hat{\tau}_{demo}|$} 
\STATE{calculate temporal scalar $\lambda = \frac{|\tau_{base}|}{|\hat{\tau}_{demo}|}$}
\FOR{each $i$ in $iters_{max}$}
\STATE{\com{// rollout $\hat{\tau}$ from state $s_0$ and actions $u$}}
\STATE{$\hat{\tau} \gets rollout(s_0, u, f_{dyn})$}
\STATE{\com{// Gradient descent on $u$ with current $C_\phi$}}
\STATE{${u}_{new} \gets {u} - \alpha.\nabla_{{u}}C_{\phi}(\lambda x, \hat{\tau}, g)$}
\ENDFOR
\STATE{\com{// Update $\phi$ based on $u_{new}$'s performance}}
\STATE{$\hat{\tau} \gets rollout(s_0, u_{new}, f_{dyn})$}
\STATE{\com{// Computes gradient through the inner loop}}
\STATE{$\phi \gets \phi - \eta.\nabla_\phi \irlloss(\hat{\tau}, \tau_{demo})$}
\ENDFOR
\ENDFOR
}
\end{algorithmic}
\caption{\strut\small Gradient-Based IRL for misaligned demonstrations}
\label{algo:cost-learning-meta-train-individual}
\vspace{-3mm}
\end{algorithm}
\subsection{Gradient-Based Model-Based IRL}
\label{sec:gradient-mb-irl}
We learn time-invariant costs through model-based IRL. In this context, we assume that we have access to a differentiable dynamics model (learnt or known), $f_{dyn}(\cdot)$ which we use at every inner update step to unroll a full trajectory and perform one action optimisation step using a learnable cost $C_{\phi}$ as shown in line 14 in Alg.~\ref{algo:cost-learning-meta-train-individual}. In each inner loop we perform few-shot action optimisation over $\mathbf{u}$ for some policy by relying on the current parameters $\phi$ (e.g. rows 10-15). We then use the produced trajectory $\hat{\tau}$ from the extracted policy to update parameters $\phi$ (lines 16-19) as part of the outer loop. The role of bi-level optimisation in this context is in connecting the gradients between the two loops, establishing a model-based cost learning solution. This results in more stable and efficient model-based optimisation~\cite{das2020model}.
\subsection{Learning time-invariant cost functions}
Learning skills for object manipulation from demonstrations can be difficult. One reason for this is that providing consistent demonstrations at a safe but sufficiently useful speed for a robot can be a tedious and often impractical solution. Furthermore, assuming that policies using the learned costs can operate at a single fixed speed can limit the utility of a given solution requiring re-taking demonstrations to scale to different speeds. To alleviate this, we propose two distinct time-invariant costs that rely on a temporal scaling term $\lambda$ that allows for extracting costs that are independent of the speed of execution.

In particular, we propose a structured cost defined using a weighted RBF network and an unstructured cost defined with an MLP with nonlinear activation. We show that when those costs utilise $\lambda$ they perform well when trained with either aligned or misaligned demonstrations.

\paragraph{Temporal Scaling with a structured cost}
Learning a cost function (used in line 14 in Alg.~\ref{algo:cost-learning-meta-train-individual}) that is linear in the parameters is often a preferred choice in the model-based IRL literature \cite{abbeel2010autonomous, englert2017inverseRL, das2020model}. Such costs, are interpretable and require less training to learn which makes them excellent for robotic applications. One successful approach for extracting useful parameters in this context is by incorporating additional structure in the cost definition, 
\begin{equation}
    C_\phi(\hat{\tau}, g) =  \mathbb{E}_{\hat{\tau}} \left[  \phi(\hat{\tau} - g)^2 \right],
    \label{eqn:cost}
\end{equation}
where $(\hat{\tau}-g)^2$ is a feature of the input trajectory $\hat{\tau}$ with respect to some goal $g$ and $\phi$ is some learnable parameterisation.
Recently, \cite{das2020model} compared a range of linear in the parameters cost functions and showed that using a time dependent weighted cost can have positive effects to successfully extracting strategies from demonstration. In particular, \cite{das2020model} showed that an RBF-based structured cost with Gaussian kernel can outperform a range of other linear costs with fixed time dependencies in the context of object placement tasks. This motivated our choice of a weighted RBF network as a structured cost. In the context of Eq.~\ref{eqn:cost}, an RBF-based structured cost can be expressed by setting $\phi = \omega \cdot e^{b(x - \mu)^{2}}$, for a fixed linearly spaced variable $x$ acting as time stamps for the desired output. Here the RBF weights $\omega$ are meta learnt. We used linearly spaced in time fixed number of kernel centres $\mu$, and $b$ is chosen to create some overlap between the neighbouring kernels. However, this as well as many common strategies assume fixed structuring of those costs \cite{abbeel2004apprenticeship, finn2016guidedirl, das2020model}. That is, once learnt, those costs cannot generalise to varied speeds of execution which limits their applicability. In this work we show how to automatically scale $C_{\phi}$ such that the induced structure can shift in time but stays the same relative to the given strategy extracted from demonstration (See Figure~\ref{fig:perturbations}). We refer to this ability as temporal scaling.

Our structured time-invariant cost function is similar to the one above and uses a linearly spaced in time fixed number of kernel centres $\mu$. In practice, this ensures that a given centre, $\mu$ will influence a specific region of a chosen trajectory (see Figure~\ref{fig:perturbations}). Shifting those centres in time but keeping them in the same region spatially would allow for a weighted representation that can adjust to different speeds of execution. Our temporally invariant learned cost is then obtained by using a temporal term $\lambda$ that is proportional to the desired speed of execution given some base speed $|\tau_{base}|$ extracted from the initial demonstrations, resulting in the following time-invariant structured cost
\begin{equation}
C_\phi(\lambda x, \hat{\tau}, g) =  \mathbb{E}_{\hat{\tau}} \left[\omega \cdot e^{b(\lambda x - \mu)^{2}} (\hat{\tau} - g)^2 \right], 
\label{eqn:structured}
\end{equation} where we define $\lambda = \frac{|\tau_{base}|}{|\hat{\tau}|}$, for some desired duration $|\hat{\tau}|$. This way the associated bandwidths get scaled proportionally to the length of $\tau_{base}$, $|\tau_{base}|$, depending on whether the desired test trajectory is faster or slower with respect to the average speed used during training (Figure~\ref{fig:perturbations}, final plot). In this context, slower trajectories relative to the training data speed would be comprised of more data points (see Figure~\ref{fig:perturbations}, first three plots). Therefore, the learned weights would be responsible for modelling more points on average, thus the bandwidth of a single RBF needs to be spread across larger number of timesteps. Then, a smaller term $\lambda < 1$ scales the learned cost to slower trajectories while a larger one $\lambda > 1$ is used to speed it up. Note that recomputing these terms is now decoupled from the reward learning process. Thus, we can recompute $\lambda$ independently for each demonstration (lines 6 to 9) and provide the information to the learnt rewards. 
\paragraph{Temporal Scaling with an unstructured cost}
Utilising structured rewards such as the RBF weighted cost function introduced in the previous subsection can be easier to train and can significantly improve interpretability of the learned costs. However, it takes an expert to build good structured cost functions. This can be time consuming but also constraining in a sense that we do not let the network uncover optimal behaviours. An alternative approach is to use an unstructured cost $C_{\phi}(.)$ which can learn a useful parameterisation that can be more effective but slower to learn. In this context, we utilise the same concept of temporal scaling ($\lambda$) where we use this as an additional input to an MLP with as few as two layers. 
Although neural network based costs come with their own limitation, a nice feature they have is that they do not necessarily require the same level of expert structure design. This makes them more flexible and with a bigger capacity to learn useful representations \cite{finn2016guidedirl, zhaoinverse}. Unlike the structured case, temporal scaling for such cost functions, however, is less intuitive and therefore more difficult to train. We use the temporal scaling term mechanism described in the previous section as part of the input to the unstructured cost where we define this cost as
\begin{equation}
    C_\phi(\lambda x, \hat{\tau}, g) =  \mathbb{E}_{\hat{\tau}} \left[ \sigma_{\phi}(\lambda x, \hat{\tau}, g) \right].
\end{equation} Here $\phi$ from Eq. 1 takes up the form of $\sigma_{\phi}(.)$ which is an MLP with a single hidden layer of size 16, parameterised by $\phi$ and a Sigmoid activation. We use the mean squared distance between $\hat{\tau}$ and $g$ as input feature and a temporal scaling term, $\lambda$ as input, resulting in input dimensionality of $|S| + 1$ where $|S|$ is the size of the state. Note that we multiply $\lambda$ by the same fixed linear spaced variable $x$ as in Eq.~\ref{eqn:structured}. This feature $\lambda x$ is indirectly connected with the locality of weights discussed in the previous subsection. We found that this cost required both more data and more inner update steps (e.g. five or more) to generate a useful feature representation which led to slower training (see Figure~\ref{fig:grid_search}). The classic MLP we benchmark against in this work assumes the exact same architecture as the one described above. However, it does not take in as input the produced timestamps $\lambda x$.

\begin{figure*}
\centering
      \includegraphics[width=\textwidth]{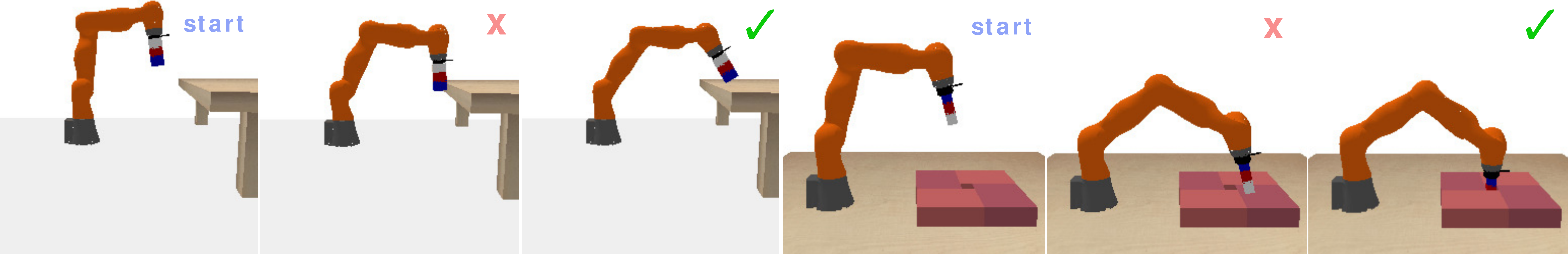}
      \caption{\small{Environment tasks are defined by varying the speed of execution and the xy location of the goals away from the base of the robot. Figure shows start configuration, failed and successful attempts.}
      }
      \label{fig:the_tasks}
\vspace{-4mm}
\end{figure*}
\section{EXPERIMENTS}
\begin{figure}[!b]
\vspace{-2mm}
\centering
      \includegraphics[width=0.48\textwidth]{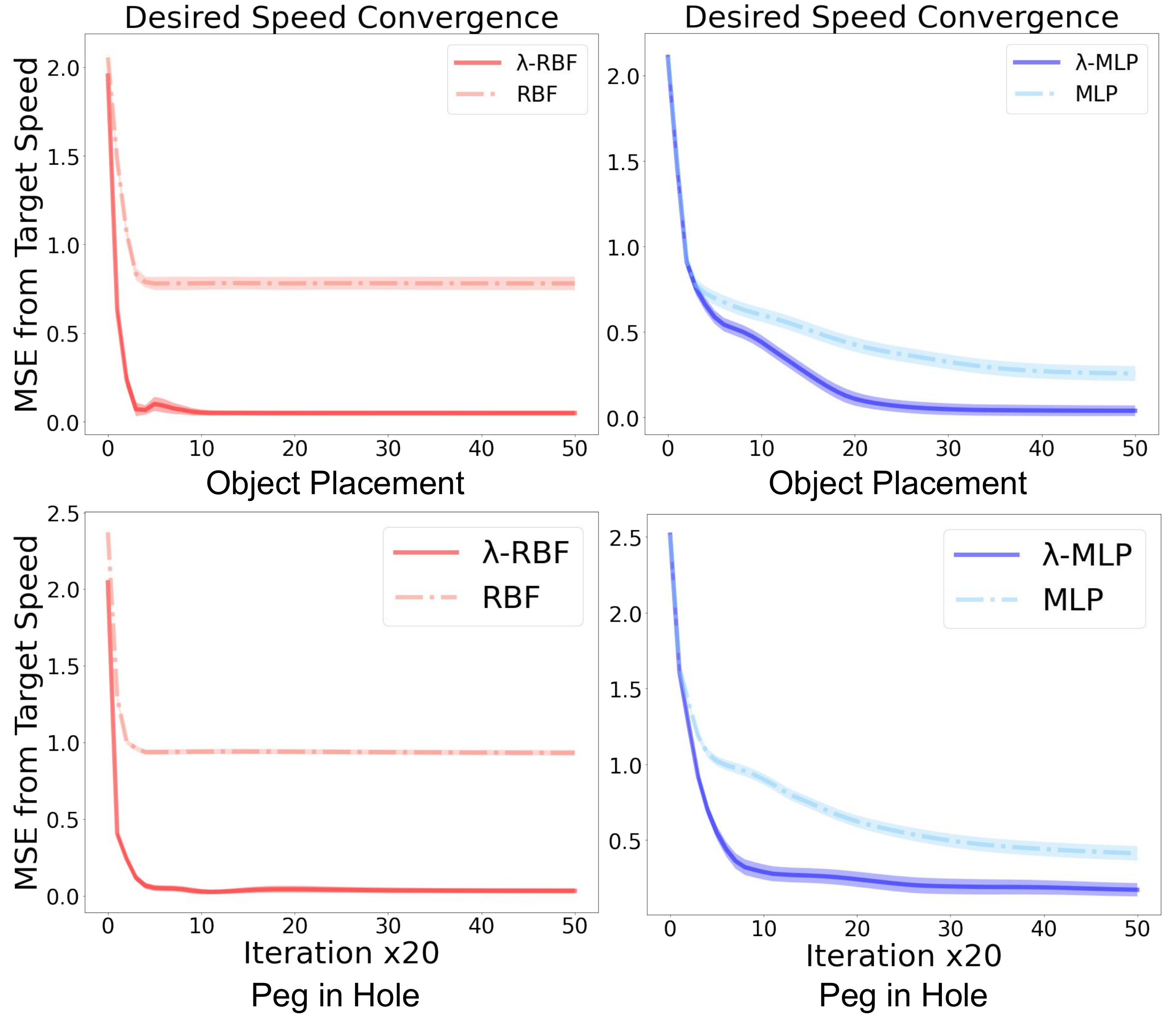}
      \caption{\small{Task convergence at train time. Temporal invariance leads to learning more flexible costs.}
      }
      \label{fig:speed-convergence}
\end{figure}
In this section we evaluate the utility of temporal scaling in the context of learning costs through model-based IRL. Our study shows that temporal scaling can help IRL generalise better with respect to violations of the temporal alignment assumptions. Further, we show that our solution can learn from multiple demonstrations that may vary in speed and targeted goals. Moreover, we show that our costs can generalise to different tasks represented in terms of different goals and various speeds of execution.
\begin{figure*}
\centering
      \includegraphics[width=\textwidth]{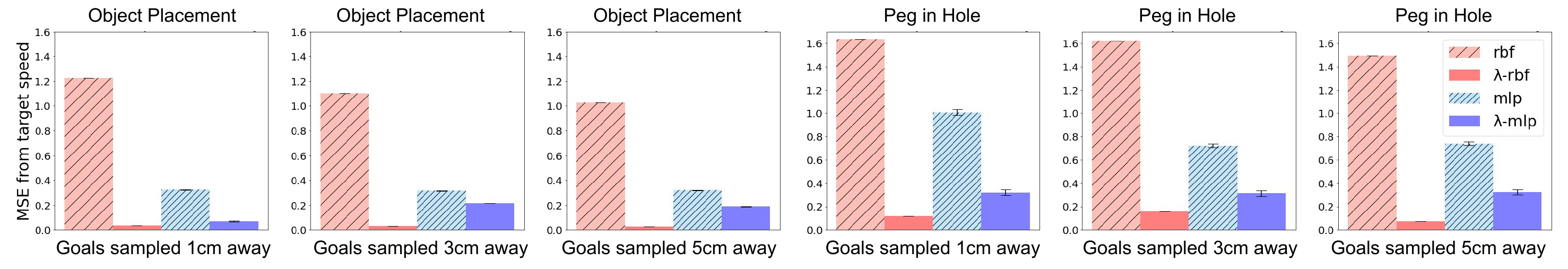}
      \caption{\small{MSE from target speeds split by distance away from centre. Each bin reports average distance on the five different speed tasks applied to 10 different goals each.}
      }
      \label{fig:achieved_speed_test}
\vspace{-6mm}
\end{figure*}
\vspace{-2mm}
\subsection{Tasks Description}
We evaluate this work in simulation on two standard benchmark tasks namely peg insertion and object placement. We build our tasks utilising the Robosuite~\cite{robosuite2020} framework but relying on PyBullet~\cite{coumans2019} as a freely distributed physics engine alternative and the fully differentiable robot model proposed in~\cite{sutanto2020encoding}. The description of the object placement on a table surface task is to follow closely a specific strategy - namely going forward first and then downwards. Otherwise, the object gets stuck at the corner of the table resulting in a failed execution (see Figure~\ref{fig:the_tasks}, first three images). The second task we consider is peg insertion (Peg/Hole) where the goal is to fit the held object in a hole situated on the table the robot is mounted on. This task differs both in terms of the robot configuration and the desired execution strategy (see Figure~\ref{fig:the_tasks}, last three images).
\subsection{Meta-training: analysis and evaluation}
The concept of meta learning has two individual meta stages, namely meta training and meta test. At meta training time we learn a reward function that can be used to perform a limited number of tasks by utilising bi-level optimisation as discussed in Section~\ref{sec:gradient-mb-irl}. This section focuses on studying the capacity of the proposed solution to learn to solve the small subset of training tasks in the context of peg-in-hole and object placement.
\paragraph{Meta-training: solving the tasks}
We report the achieved performance of the different costs on the considered training tasks when training with temporally misaligned demonstrations against different goals in Table~\ref{tabl:train_time_table}. All costs were able to solve the tasks in both environments. The structured RBF cost achieved better performance suggesting they might be overfitting to the training data. The unstructured non-temporally scaled cost achieved worse performance in terms of successfully completing both tasks suggesting that it struggles to perform well on the different tasks due to their variability in terms of speed and spatial location.
\begin{table}
\centering
		\resizebox{0.48\textwidth}{!}{
			\begin{tabular}[b]{p{1.0 cm} c c c c c c c c}
				\toprule 
				\bf Env. & \bf RBF \cite{das2020model} & \bf $\lambda$-RBF & \bf MLP & \bf $\lambda$-MLP \\ 
				& Mean (Std) & Mean (Std) & Mean (Std) & Mean (Std) \\
				\midrule
				Placing & \textbf{0.4cm} (0.0) & 0.5cm (0.0) & 3.1cm (0.0) & 0.9 (0.0) \\
                PiH & 100\% (0.0) & 100\% (0.0) & 100\% (0.0) & 100\% (0.0) \\
				\bottomrule
			\end{tabular}
		}
		\caption{\small Records performance in terms of mean squared distance in the placing environment and in terms of successful insertions for the peg in hole (PiH) environment. Train time table.}
		\label{tabl:train_time_table}
		\vspace{-6mm}
\end{table}
\paragraph{Meta-training setup}
\label{sec:meta-train-setup}
We evaluate the ability of our solution to learn rewards that can generalise to different speeds and that can be applied to different goals. To achieve this we consider learning rewards in three different contexts: a) we consider learning costs from perfectly time-aligned demonstrations performed with respect to four different goals which were sampled uniformly within 1cm range from some chosen centre; b) we fix the goal and consider learning costs from four time misaligned demonstrations that were split in two: demonstrations that took roughly 3 seconds to execute (within a range of 2.8 and 3.2 seconds); and demonstrations that took roughly 5 seconds to execute (between 4.8 and 5.2 seconds); and finally c) a mix of a) and b), namely we use 12 time misaligned demonstrations provided for different goals.
\paragraph{Scaling to misaligned demonstrations: relaxing the temporal alignment assumption}
Figure~\ref{fig:meta-train} summarises the achieved desired speeds on the meta training task. First column illustrates the achieved performance in a classic IRL setting (i.e. context `a` from above). We evaluate our solution on its ability to achieve desired speeds. We provide additional analysis including studying the ability of the solution to solve the desired spatial tasks in the Appendix Table~\ref{tabl:train_time_table}. As expected, our proposed solutions performed on-par with their non-temporally scaled alternatives. This, however, changes when we include a drastically varied speed of execution as depicted in the second (context `b`) and third (context `c`) columns. Learning structured temporally scaled RBF-based costs ($\lambda$-RBF) performed best on all tasks followed closely by the temporally varied MLP ($\lambda$-MLP). Relying on a simple unstructured network did not perform as well as either of the temporally varied variants but performed better than the structured RBF on the misaligned demonstrations. This suggests the ability of unstructured costs to pick up temporally relevant representations that are not necessarily easy to detect or model.
\paragraph{Interpreting the achieved speed convergence: comparing the different costs}
Figure~\ref{fig:speed-convergence} illustrates the achieved distance from target speeds. The results show that all costs converge very quickly and have very low variance. This is due to combining MPC with meta learning \cite{das2020model}. However, reward functions that do not have access to temporal scaling do not have enough flexibility which prevents them from reaching a good solution. Instead, the baseline cost functions learn a reward function that achieves an average speed over all demonstrations as they do not have the flexibility to disentangle the duration parameter.
Both unstructured costs performed slightly worse than their structured counter parts - we attribute this to the small number of data points considered (3 to 12 demos) and hypothesise this gap can be closed with the addition of more training data.
\subsection{Meta-testing: Transfer to new tasks}

Extracting policies from learnt temporally invariant costs can lead to flexible and more practical solutions. We evaluate this statement next by directly comparing our time-invariant costs to their non-temporally scaled versions and against using standard data alignment. Our results show that time invariance improves the applicability of the learnt costs to different speeds and task settings. 
\paragraph{Experimental Details}
We evaluate the ability of the different rewards to transfer to new tasks. We trained with 3 different seeds and using the misaligned demonstrations set up from Section~\ref{sec:meta-train-setup}. Each demo goal was sampled uniformly within 1cm away from a uniform training data distribution. The ability to scale to different speeds is evaluated on five different and never seen before speeds per environment. Specifically, we vary the tasks speeds from a very fast execution to a very slow one. The speeds we used started from task execution in 2 seconds to task execution in 6 seconds. We sampled a total of 30 goals per considered speed and split those goals into three individual bins. Each goal bin has 10 goals and is comprised of goals sampled some distance away (1, 3 and 5cm) from the goal distribution centre. None of the test goals and speeds have been seen at train time, however some of them (i.e. 1cm) are sampled from the same goal distribution as the training data. The rest are out of goal distribution samples (OOD). The overall pool of test tasks was thus comprised of both in and out of distribution examples. We use a fixed control frequency of 5Hz for running our policy in all cases. In total, we test against 150 independent evaluation runs. We used one update step for policies trained using the structured costs and five for the unstructured.\vspace{-1mm}
\subsubsection{Studying the overall performance}
\begin{figure*}[h!]
      \centering
      \includegraphics[width=\textwidth]{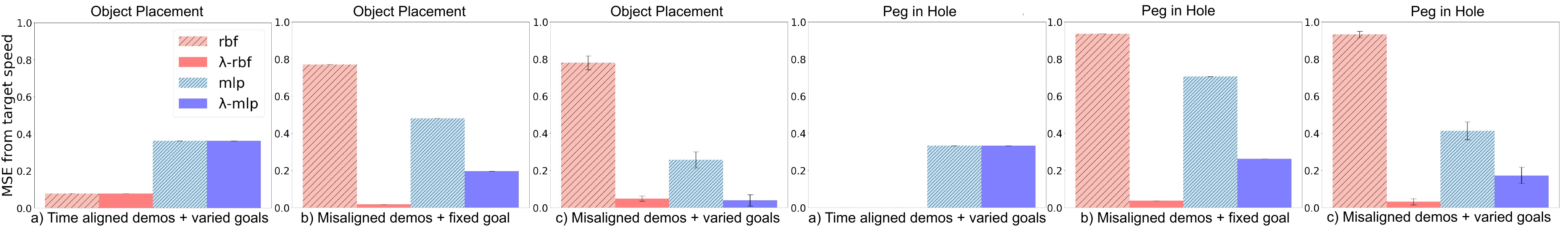}
      \caption{\small{MSE from target speeds split when trained according to three contexts. a) learning from perfectly time-aligned demonstrations and varied goals; b) learning from fixed goals but temporally misaligned demonstrations; and c) learning from temporally misaligned demonstrations for varied goals. Comparison for all costs against both object placement and peg in hole environments. Overall, temporally scaled costs can cope with the introduced challenges better than the non-temporally scaled costs. Lower is better.}}
      \label{fig:meta-train} 
      \vspace{-4mm}
\end{figure*}
We begin by studying the overall performance of the considered solutions. We report the average achieved results in Table~\ref{tabl:test_time_table}. We compare against a state-of-the-art RBF-based reward, recently introduced in \cite{das2020model} and an MLP based one. We also compare against aligning demonstrations with dynamic time warping (DTW) before training with \cite{das2020model} too. All baselines relied on meta learning similar to us (details in Appendix~\ref{sec:baseline-details}). Overall, our proposed solution significantly outperformed the alternatives. $\lambda$-MLP required more data and inner update steps making it slower to train. Our results indicate that at times it had higher variance than the RBF-based costs too. However, it required less expert knowledge and picked up useful features directly from data. The structured $\lambda$-RBF trained much faster and was far more sample efficient but performed slightly worse. In contrast, the baselines did not do well albeit their performance on the training data (Table~\ref{tabl:train_time_table}). Aligning with DTW was very sensitive to the reference demonstration which resulted in a slightly better performance than \cite{das2020model} on the Peg/Hole tasks and worse for the placing ones.
\begin{table}[H]
\centering
		\resizebox{0.48\textwidth}{!}{
			\begin{tabular}[b]{p{1.0 cm} c c c c c c c c c}
				\toprule 
				\bf Env. & \bf RBF \cite{das2020model} & \bf DTW-RBF & \bf $\lambda$-RBF & \bf MLP & \bf $\lambda$-MLP \\ 
				& Mean (Std) & Mean (Std) & Mean (Std) & Mean (Std) & Mean (Std) \\
				\midrule
				Placing & 4.08cm (0.47) & 5.88cm (0.25) & 1.08cm (0.59) & 4.10cm (0.02) & \textbf{0.96cm} (0.08) \\
                Peg/Hole & 22.22\% (0.31) & 25.1\% (0.83) & 59.33\% (0.63) & 32.45\% (8.45) & \textbf{70.22\%} (0.89) \\
				\bottomrule
			\end{tabular}
		}
		\caption{\small Records performance in terms of mean squared distance in the placing environment and in terms of successful insertions for the peg in hole (PiH) environment. Test time table.}
		\label{tabl:test_time_table}
		\vspace{-5mm}
\end{table}
\subsubsection{Evaluating out of distribution performance}
A key benefit of learning-to-learn based solutions is their ability to scale to new tasks. Here, we differentiate between two types of never seen before tasks: ones that vary according to speed of execution, and ones that vary according to a target goal and its distance from the task distribution centre. Our findings suggest that overall $\lambda$-RBF performs better against desired speed of execution and against goals closer to the distribution centre. The unstructured $\lambda$-MLP cost is better at OOD tasks suggesting its ability to learn more general features than the expert-structured ones.
\paragraph{Generalising to different speeds}
Figure~\ref{fig:achieved_speed_test} compares the achieved speeds on the considered tasks. The comparison is conducted across a sequence of desired episode lengths using the temporally scaled and the non-temporally scaled costs. Generally, the time-invariant costs  achieved similar velocities to the target ones while the non-temporally scaled reward had over-fitted to the training data. Achieving the desired speed is important to ensure a successful completion of the task. In practice, failing to scale to the appropriate speed might result in a failure to complete the task or worse, reach it sooner than anticipated and continue pushing towards e.g. a table for the remainder of the predetermined time which can be damaging to the robot itself. Unlike the basic RBF cost, the basic MLP cost is somewhat susceptible to speeds due to its independence from the length of the trajectories. This means that given enough data and training time the MLP based cost might be able to extract useful temporally invariant features as suggested by the last figure in Figure~\ref{fig:meta-train} and the results reported in Figure~\ref{fig:achieved_speed_test}.
\paragraph{Generalising to different goals}
We extend the experimental results reported in the previous section and disentangle the reported values from Table~\ref{tabl:test_time_table}. An ideal solution would achieve similar test set performance to the training set performance. Similarly, a less flexible cost would result in drop in performance the further away it gets from the training distribution centre. Table~\ref{tabl:test_time_disentangled_table} reports our findings for the Peg/Hole environment. The performance of the temporally invariant costs is significantly better than the alternatives. $\lambda$-RBF performs best on tasks that are more similar to the training data suggesting that the proposed structure is good for in-distribution problems. However, its performance drops drastically on both OOD tasks. In contrast, the good performance of $\lambda$-MLP to OOD tasks indicates its ability to generalise better to out of distribution tasks possibly due to learning more general feature representations. We dedicate the performance of the baselines to their inability to learn consistently from misaligned demonstrations and the compliance of the environment. Using DTW can fix this to some extent but is not sufficient as the learnt reward itself is not time-invariant.

\begin{table}[H]
\centering
\resizebox{0.48\textwidth}{!}{
			\begin{tabular}[b]{p{1.0 cm} c c c c c c c c}
				\toprule 
				\bf Task & \bf RBF \cite{das2020model} & \bf DTW-RBF & \bf $\lambda$-RBF & \bf MLP & \bf $\lambda$-MLP \\ 
				& Mean (Std) & Mean (Std) & Mean (Std) & Mean (Std) & Mean (Std) \\
				\midrule
				1cm & 26.0\% (0.0) & 44.0\% (0.0) &  \textbf{86.0\%} (0.0) & 28.0 (12.73) & 78.0\% (9.43) \\
                3cm & 10.67\% (0.94) & 15.3\% (2.49) & 44.0\% (0.0) & 30.67\% (4.46) & \textbf{74.0\%} (3.77) \\
                5cm & 39.0\% (0.0) & 16.0\% (0.0) & 48.0\% (1.89) & 38.67\% (8.15) & \textbf{58.67\%} (0.94) \\
                \midrule
                Avg. & 22.22\% (0.31) & 25.11\% (0.83) & 59.33\% (0.63) & 32.45\% (8.45) & \textbf{70.22\%} (0.89) \\
				\bottomrule
			\end{tabular}
			}
		\caption{\small Percent successful insertions for the peg in hole (PiH). Structured cost is more accurate on closer samples while the unstructured cost performs better on out-of-distribution samples.}
		\label{tabl:test_time_disentangled_table}
\end{table} \vspace{-5mm}
\paragraph{Generalising to different goals (Placing)}
We evaluate the ability of the considered rewards to solve the different placing tasks too. We provide the associated results in Table~\ref{tabl:test_time_table_placing}. Similar performance can be consistently seen here too except the accuracy in performance of all baselines drops linearly with the complexity of the task. We dedicate this to the smaller dependency on the environment's compliance in solving the task.
\begin{table}[H]
\centering
\resizebox{0.48\textwidth}{!}{
			\begin{tabular}[b]{p{1.0 cm} c c c c c c c c c}
				\toprule 
				\bf Task & \bf RBF \cite{das2020model} & \bf DTW-RBF & \bf $\lambda$-RBF & \bf MLP & \bf $\lambda$-MLP \\ 
				& Mean (Std) & Mean (Std) & Mean (Std) & Mean (Std) & Mean (Std) \\
				\midrule
				1cm & 3.49cm (2.87) & 5.58cm (3.45) & \textbf{0.35cm} (0.31) & 4.08cm (2.30) & 0.87cm (0.30)\\
                3cm & 4.13cm (2.73) & 5.85cm (3.32) & 1.09cm (0.20) & 4.10cm (2.24) & \textbf{0.94cm} (0.55) \\
                5cm & 4.63cm (2.63) & 6.20cm (3.16) & 1.79cm (0.16) & 4.13cm (2.20) & \textbf{1.07cm} (0.71)\\
                \midrule
                Avg. & 4.08cm (0.47) & 5.88cm (0.25) & 1.08cm (0.59) & 4.1cm (0.02) & \textbf{0.96cm} (0.08) \\
				\bottomrule
			\end{tabular}
			}
		\caption{\small Records performance in terms of mean squared distance in the placing environment and in terms of successful insertions for the object placement environment. Structured cost is more accurate on closer samples while the unstructured cost performs better on out-of-distribution samples.}
		\label{tabl:test_time_table_placing} 
\end{table}
\vspace{-6mm}
\section{Conclusion}
We introduce a method for learning time-invariant costs through model-based inverse reinforcement learning. We build upon recent advances in learning-to-learn model-based IRL \cite{das2020model} and evaluate our approach on two sets of basic but common manipulation tasks, that of object placement and peg insertion. We propose two different time-invariant costs - a structured and an unstructured one. We study their capacity to generalise to different execution durations while remaining robust to misaligned demonstrations. Our analysis highlights the ability of a structured cost to enable fast and successful acquisition of a demonstrated strategy while remaining intuitive to understand. We observe that employing an unstructured cost, even if it requires five times longer training,  maintains not only the time-invariant property of our definition, but also shows consistent performance across out of distribution tasks unlike its structured variant. Directions for future work include extensions to handling multiple control frequencies within a single execution. In addition, we expect this to enable the ability of the unstructured cost to fuse different sensors at a time.
\section*{APPENDIX}
\section{Evaluation Details}\label{appendix:relative_dist}
We evaluate the baseline and learned cost functions similar to \cite{das2020model} by comparing our state from the last step of a planned trajectory we optimise with the goal. The planned trajectory is extracted using Algorithm \ref{algo:planning}.
\vspace{-3mm}
\begin{algorithm}[H]
\begin{algorithmic}[1]
\footnotesize{
 \renewcommand{\algorithmicrequire}{\textbf{Input:}}
 \renewcommand{\algorithmicensure}{\textbf{Output:}}
\REQUIRE cost $C_\phi$, planning horizon $T$, the forward dynamics model $f_\text{dyn}$, goal $g$ and an initial state $s_0 = [\theta_0, z_0]$, $\theta_t$ denotes the joint vector at time $t$.
\ENSURE  $\tilde z, \tilde \theta$
\vspace{2mm}
\STATE{Initialize ${u}_{\text{init}, t} = 0, \forall t=1, \dots, T$ and $\lambda = \frac{|\tau_{base}|}{T}$ with a fixed linearly spaced variable $x$, for $|x| = T$}
\FOR{each $i$ in $iters_{max}$}
\STATE{\com{$//$ MPC-based rollout using the initial actions}}
\STATE{$\hat \theta_0 = \theta_0$}
\STATE{$\hat \tau = \{\hat z_0\}$}
\FOR{\hspace{1mm}$t \gets 1:T$}
\STATE{$\hat s_{t-1} = [\hat z_{t=1}, \hat \theta_{t-1}, \hat{\dot{\theta}}_{t-1}]^T$}
\STATE{$\hat z_t, \hat \theta_t, \hat{\dot{\theta}}_{t} = f_\text{dyn}(\hat s_{t-1}, u_{\text{init}, t-1})$}
\STATE{$\hat \tau \gets \hat \tau \cup \hat{z}_t$}
\ENDFOR
\STATE{\com{$//$Action optimization}}
\STATE{${u}_\text{opt} \gets {u_\text{init}} - \alpha.\nabla_{{u}}C(\lambda x, \hat \tau, g)$}
\ENDFOR
\STATE{\com{$//$Get planned trajectory by rolling out ${u}_\text{opt}$}}
\STATE{$\tilde z_0 = z_0, \tilde \theta_0 = \theta_0, \tilde{\dot{\theta}}_0 = \dot{\theta}_0$}
\FOR{$t \gets 1:T$}
\STATE{$\tilde z_t, \tilde \theta_t = f_\text{dyn}([\tilde z_{t-1}, \tilde \theta_{t-1}, \tilde{\dot{\theta}}_{t-1}], u_{\text{opt}, t-1})$}
\ENDFOR
}
\end{algorithmic}
\caption{\strut\small Trajectory planning using given Cost}
\label{algo:planning}
\end{algorithm}
\vspace{-8mm}
\section{Baseline implementation details}
\label{sec:baseline-details}
\paragraph{RBF}
In this work we compare against a recent model-based IRL approach  \cite{das2020model} that relies on meta-learning similar to us. We used their publicly available code. 
\paragraph{DTW-RBF}
We adapt \cite{das2020model} to handle temporally misaligned demonstrations by using dynamic time warping (DTW). We refer to this as DTW-RBF. We used 'dtwalign', a publicly available python package. We found DTW to be highly sensitive to the selected reference demonstration for alignment. We used the slow-paced execution (i.e. 5.2 seconds) with cross validation over all available demonstrations.

\paragraph{MLP} We used the same MLP-based cost structure as for our $\lambda$-MLP except we took out the $\lambda x$ term. We used 1 16 dimensional MLP with a sigmoid activation, meta learning rate of 1e-3 and a task learning rate of 3e-1. We trained using 1000 epochs and 5 inner update steps.

\section*{ACKNOWLEDGMENT}

The authors would like to thank S. Schaal and the RAD group for valuable comments and feedback on earlier drafts of this work.

\bibliographystyle{IEEEtran} 
\bibliography{IEEEexample}

\end{document}